\documentclass[12pt,a4paper]{article}
\usepackage[top=4cm, bottom=4cm, left=3cm, right=2.5cm]{geometry}
\usepackage[utf8]{inputenc}
\usepackage{amsmath}
\usepackage{amsfonts}
\usepackage{amssymb}
\usepackage{amsmath}
\usepackage{verbatim}
\usepackage{mathrsfs}
\usepackage{mathtools}
\usepackage{xcolor}
\usepackage{titling}

\usepackage{setspace}
\usepackage{amsthm}
\usepackage{pythonhighlight}

\newtheorem{definition}{Definition}
\newtheorem{theorem}{Theorem}

\usepackage[backend=bibtex, style=authoryear, useprefix]{biblatex}
\bibliography{bib.bib} 
\DeclareFieldFormat[article]{pages}{#1}%
\DeclareFieldFormat[incollection]{pages}{#1}%
\DeclareFieldFormat{postnote}{#1}
\DeclareFieldFormat{multipostnote}{#1}

\usepackage{hyperref}

\author{Sven Neth \\ \small{University of California, Berkeley} \\ \small{nethsven@berkeley.edu}}
\title{A Dilemma for Solomonoff Prediction}

\begin{document}

\begin{titlingpage}
    \maketitle
    \begin{abstract}
The framework of \emph{Solomonoff prediction} assigns prior probability to hypotheses inversely proportional to their Kolmogorov complexity. There are two well-known problems. First, the Solomonoff prior is relative to a choice of Universal Turing machine. Second, the Solomonoff prior is not computable. However, there are responses to both problems. Different Solomonoff priors \emph{converge} with more and more data. Further, there are \emph{computable approximations} to the Solomonoff prior. I argue that there is a tension between these two responses. This is because computable approximations to Solomonoff prediction do \emph{not} always converge.
    \end{abstract}
    \textbf{Acknowledgements}: I would like to thank Lara Buchak, John MacFarlane, Thomas Icard and two anonymous referees for  helpful comments on earlier drafts. I presented this material at the 27th Biennial Meeting of the Philosophy of Science Association in November 2021 and would like to thank the audience for asking good questions. Further thanks to Snow Zhang, Kshitij Kulkarni and Reid Dale for helpful discussion and J\"urgen Neth for comments on the final manuscript. Special thanks to the editors who were very helpful and accommodating when I faced some unforeseen challenges in finishing this paper. During research, I was supported by a Global Priorities Fellowship by the Forethought Foundation.
\end{titlingpage}

\section{Introduction}

We are often interested in how to make predictions on the basis of observed data. This question is at the heart of scientific inference and statistics. It is also important for the project of building artificial intelligence which can make inferences from observed data and act accordingly. Thus, there are many good reasons to be concerned about the right framework for predictive inference.

One way to tackle this question is the \emph{Bayesian approach}, which uses a prior probability distribution over all relevant hypotheses and then updates this prior by conditionalization on the observed data \parencite{Earman1992}. The resulting posterior distribution can be used to make predictions and guide action. The Bayesian approach gives us a unified framework to think about predictive inference and has been successfully applied across many fields, from astronomy to finance. However, the Bayesian approach requires us to start with a prior probability distribution over all relevant hypotheses. How should we select such a prior probability distribution? This is the \emph{problem of the priors}.

A natural response to the problem of the priors is to say that we should assign higher prior probability to \emph{simpler} hypotheses. This idea is often known as `Ockham's razor' and seems intuitively appealing to many people. However, how do we measure the simplicity of hypotheses? A possible answer to this question is provided by the framework of \emph{Solomonoff prediction}, which formalizes the simplicity of hypotheses using tools from algorithmic information theory \parencite{Solomonoff1964, Hutter2007, Sterkenburg2016, Li2019}. The \emph{Solomonoff prior} assigns higher probability to hypotheses which are simpler in this sense. Since the Solomonoff prior is defined for a very broad range of hypotheses, it provides a very general response to the problem of the priors. Moreover, proponents of Solomonoff prediction argue that the Solomonoff prior is an `objective' and `universal' prior. Thus, the framework of Solomonoff prediction potentially sheds light on the foundations of scientific inference, the problem of induction and our prospects for building `Universal Artificial Intelligence' \parencite{Hutter2004}.

There are two well-known problems for Solomonoff prediction. First, the Solomonoff prior is relative to a choice of Universal Turing machine, which means that different choices of Universal Turing machine lead to different priors and different predictions. It is natural to worry that this undermines the ambition of Solomonoff prediction to provide an `objective' and `universal' prior. Second, the Solomonoff prior is not computable, which means that no scientist or AI system could actually use the Solomonoff prior to make predictions.  

There are well-known responses to both objections. While it is true that the Solomonoff prior is relative to a choice of Universal Turing machine, it can be shown that different Solomonoff priors \emph{converge} with more and more data (in a sense which will be made precise below). Further, while the Solomonoff prior is not computable, there are \emph{computable approximations} to it. 

I argue that there is a deep tension between these two responses. This is because different computable approximations to Solomonoff prediction do \emph{not} always converge. Therefore, if we care about universal convergence, computable approximations to Solomonoff prediction do not give us what we want. Thus, proponents of Solomonoff prediction face a pressing dilemma. Either they have to give up universal convergence, which leads to problems of language dependence and subjectivity. Or they have to accept that Solomonoff prediction is essentially uncomputable and so cannot be of any help to guide the inferences of human and artificial agents. Therefore, Solomonoff prediction does \emph{not} solve the problem of finding a universal prior probability distribution which can be used as a foundation for scientific inference and artificial intelligence.

\section{Solomonoff Prediction}

I start by giving a brief introduction to Solomonoff prediction \parencite{Solomonoff1964, Hutter2007, Sterkenburg2016, Li2019}.\footnote{For more discussion, see \cite{Ortner2009, Rathmanner2011, Vallinder2012, Chater2013, Icard2017, Sterkenburg2018thesis}.}

Suppose you are given this initial segment of a binary string:
\begin{equation*}
00000000...
\end{equation*}
Given this initial segment, what is your prediction for the next bit? 

In a Bayesian framework, we can answer this question by consulting a \emph{prior probability measure} over the set of all binary strings. To make this answer precise, we first need to introduce some notation. Let $\mathcal{B}^\infty$ be the set of all infinite binary strings and $\mathcal{B}^*$ be the set of all finite binary strings. If $x \in \mathcal{B}^*$ and $y \in \mathcal{B}^* \cup \mathcal{B}^\infty$, we write $xy$ to denote the \emph{concatenation} of $x$ and $y$, the (finite or infinite) binary string which starts with $x$ and continues with $y$. We say that $x$ is a (proper) \emph{prefix} of $y$ if $y = xz$ for some string $z$ (and $z$ is not the empty string).

At first, we focus on a particular kind of set of infinite binary strings:
\begin{definition}
For every $x \in \mathcal{B}^*$, the \emph{cylinder} $\Gamma_x \subseteq \mathcal{B}^\infty$ is defined by $\Gamma_x = \{ x\omega : \omega \in \mathcal{B}^\infty \}$ \parencite[265]{Li2019}.
\end{definition}
Intuitively, a cylinder is a set of binary strings which begin with the same string and then diverge. For example, $\Gamma_1 =  \{ 1\omega : \omega \in \mathcal{B}^\infty \}$ is the set of all binary strings which begin with $1$. We write $\epsilon$ for the empty string. Therefore,  $\Gamma_{\epsilon}$ is the set of all binary strings which begin with the empty string, which is just the set of all binary strings. We write $\mathfrak{C}$ for the set of all cylinders.

With this framework in place, we can define a probability measure as follows. First, we define:
\begin{definition}
A \emph{pre-measure} is a function $p : \mathfrak{C} \to [0,1]$ such that
\begin{enumerate}
\item $p(\Gamma_{\epsilon}) = 1,$
\item $p(\Gamma_x) = p(\Gamma_{x0}) + p(\Gamma_{x1})$ for all $x \in \mathcal{B}^*$.
\end{enumerate}
\end{definition}
Intuitively, a pre-measure assigns probabilities to all cylinder sets. Once we have defined probabilities for all cylinder sets, we can extend our assignment of probabilities to more complicated sets. Let $\mathfrak{F}$ be the result of closing $\mathfrak{C}$ under complementation and countable union. Thus, $\mathfrak{F}$ is a $\sigma$-algebra. By Carath{\'e}odory's Extension theorem, every pre-measure  $p : \mathfrak{C} \to [0,1]$ determines a unique probability measure $p:  \mathfrak{F} \to [0,1]$ which satisfies the standard Kolmogorov axioms.\footnote{\textcite[64]{Sterkenburg2018thesis} sketches a more detailed version of this argument. A similar application of Carath{\'e}odory's Extension theorem is discussed by \textcite[61]{Earman1992}.} In light of this, we will abuse notation in what follows and sometimes refer to a pre-measure $p : \mathfrak{C} \to [0,1]$ as a probability measure. If $x \in \mathcal{B}^*$, we will often write $p(x)$ to abbreviate $p(\Gamma_x)$.

Now the basic idea of Solomonoff prediction is that we should assign higher prior probability to \emph{simpler} binary strings. However, what do we mean by `simplicity' or `complexity'? We can formalize the complexity of a string as its \emph{Kolmogorov complexity}: the length of the shortest program in some universal programming language which outputs that string. We can model a universal programming language as a monotone Universal Turing machine $U$ \parencite[303]{Li2019}. A monotone Universal Turing Machine has a one-way read-only input tape and a one-way write-only output tape. The input tape contains a binary string which is the \emph{program} to be executed, and the output tape contains a binary string which is the \emph{output}. The Turing machine must further be \emph{universal}, which means that it can emulate any computable function. Finally, to say that the Turing machine is \emph{monotone} means that the output tape is write-only, so the machine cannot edit its previous outputs.\footnote{The focus on monotone machines is to ensure, via Kraft's inequality, that the sum in (\ref{solprior}) is less than or equal to one \parencite[275]{Li2019}. See also Definition 2 in \cite{Wood2013}.}

Then, we define the \emph{Solomonoff prior}, which assigns prior `probability' to binary strings inversely proportional to their Kolmogorov complexity. For every finite binary string $b \in \mathcal{B}^*$, we have:
\begin{equation}\label{solprior}
\lambda_U(b) = \sum_{\rho \in D_{U, b}} 2^{- \ell(\rho)},
\end{equation}
where $D_{U, b}$ is the set of minimal programs which lead $U$ to output a string starting with $b$ and $\ell(\rho)$ is the length of program $\rho$. To say that $D_{U, b}$ is the set of minimal programs which lead $U$ to output a string starting with $b$ means that (i) upon reading any program in $D_{U, b}$, $U$ will output a string starting with $b$ and (ii) no proper prefix of any program in $D_{U, b}$ leads $U$ to output a string starting with $b$.\footnote{See \cite[307]{Li2019}, \cite[466]{Sterkenburg2016}, Definition 5 in \cite{Wood2013}.} As a rough heuristic, we can think of $\lambda_U(b)$ as the `probability' of producing the string $b$ by feeding random bits to the Universal Turing Machine $U$ on its input tape. (As we will see in a moment, the Solomonoff prior is not a probability measure, so this is not quite correct.)

As a simple example, consider a binary string which consists in a very long sequence of zeros:
\begin{equation*}
000000000...
\end{equation*}
Here $D_{U, b}$ is the set of minimal programs which output a very long sequence of zeros. In Python, one of these might be the following program $\rho$:\footnote{Both here and below, I do \emph{not} claim that these are actually minimal programs but merely use them as simple toy examples.}
\begin{python}
while True:
    print(0)
\end{python}
In this example, $\ell(\rho)$ is the Kolmogorov complexity of our string since it is the length of one of the minimal programs which outputs our string. To find the Solomonoff prior of our string, we start by computing $2^{-\ell(\rho)}$. However, there might be more than one minimal program which outputs our string. To take this into account, we take the sum over \emph{all} such minimal programs, resulting in formula (\ref{solprior}). As this example shows, there are two assumptions build into this framework. First, strings which are produced by \emph{simpler} programs should get a higher prior probability. Second, strings which are produced by \emph{more} programs should get a higher prior probability.

Each Solomonoff prior $\lambda_U(\cdot)$ induces a Solomonoff predictor, which we can write as follows for every $x \in \mathcal{B}^*$:
\begin{equation}\label{solpredictor}
\lambda_U({x1}\mid {x}) = \frac{\lambda_U({x1})}{\lambda_U({x})},
\lambda_U({x0}\mid {x}) = 1- \lambda_U({x1}\mid {x}).
\end{equation}

Intuitively, $\lambda_U({x1}\mid {x})$ tells us the probability that the next bit is $1$ given that we observed a string starting with $x$. So if we fix a Universal Turing machine $U$, this answers our earlier question what we should predict about the next bit after seeing some initial sequence. The hope is that we can encode all real-world inference problems as problems about predicting the next bit of a binary sequence. If this is possible, we can use the Solomonoff predictor to predict any kind of real-word event: the probability that the sun will rise tomorrow, the probability that the stock market will go up next month and so on.\footnote{In any concrete application, our predictions will depend not only on the Solomonoff prior, but also on how we encode a given real-world inference problem as a binary sequence. There are many different ways to represent (say) the state of the stock market as a binary sequence. Thus, there is a worry about language dependence here. However, I will bracket this worry, as it turns out that there is another more direct worry about language dependence, to be discussed in section (\ref{relativity-convergence}) below.}

As suggested above, the Solomonoff prior is \emph{not} a pre-measure on $\mathfrak{C}$. In particular, we only have
\begin{enumerate}
\item $\lambda_U({\epsilon}) \leq 1,$
\item $\lambda_U(x) \geq  \lambda_U({x0}) + \lambda_U({x1})$
\end{enumerate}
for $x \in \mathcal{B}^*$. However, sometimes these inequalities will be strict \parencite[Lemma 15]{Wood2013}. Therefore, the Solomonoff prior is only a \emph{semi-measure}, which we can think of as a `defective' probability measure. This is a problem, because there are good reasons to think that rationality requires adherence to the axioms of probability. There are \emph{dutch book arguments}, going back to \textcite{DeFinetti1937}, which show that probabilistically incoherent credences lead agents to accept a sequence of bets which are jointly guaranteed to yield a sure loss. Further, there are \emph{accuracy dominance arguments} which show that probabilistically incoherent credences are guaranteed to be less accurate than some probabilistically coherent credences.\footnote{Standard accuracy arguments are formulated in a setting with a finite algebra of events \parencite{Predd2009, Pettigrew2016}. However, there are extensions of these arguments to infinite algebras  \parencite{KelleyForthcoming}.} Therefore, from a Bayesian point of view, the Solomonoff prior is arguably a non-starter if it does not satisfy the axioms of probability. Call this the \emph{semi-measure problem}.

To fix this problem, we can define the \emph{normalized Solomonoff prior} $\Lambda_U$ as follows \parencite[308]{Li2019}. We have $\Lambda_U(\epsilon) = 1$ and for every $x \in \mathcal{B}^*$, we recursively define:\begin{equation}\label{normsolprior}
\Lambda_U({x1}) = \Lambda_U(x)\left( \frac{\lambda_U({x1})}{\lambda_U({x0}) + \lambda_U({x1})} \right), \Lambda_U({x0}) = 1 - \Lambda_U({x1}).
\end{equation}
\noindent $\Lambda_U$ is a pre-measure on $\mathfrak{C}$ and so determines a unique probability measure on $\mathfrak{F}$.\footnote{There are different ways to normalize $\lambda_U$ which is a potential source of subjectivity and arbitrariness. I will not pursue this line of criticism here. \textcite[Section 4.7]{Li2019} provide a great historical overview of the different approaches to the semi-measure problem by Solomonoff, Levin and others}

Alternatively, we can interpret the (unnormalized) Solomonoff prior $\lambda_U$ as a probability measure on the set of infinite and \emph{finite} binary strings \parencite[641]{Sterkenburg2019}. From this perspective, cases in which  $\lambda_U(x) >  \lambda_U({x0}) + \lambda_U({x1})$ represent a situation in which $\lambda_U$ assigns positive probability to the possibility that the binary string ends after the initial segment $x$. 

Does it matter which of these strategies we pick? It turns out that there is an interesting connection between normalization and the approximation reply to be discussed below. In particular, normalizing the Solomonoff prior makes it \emph{harder} to maintain the approximation reply. But the point of this paper is that there is a tension between the approximation reply and the convergence reply, and this tension will arise no matter how we deal with the semi-measure problem. Therefore, my main argument is not much affected by this choice

\section{Relativity and Convergence}\label{relativity-convergence}

We have defined the Solomonoff prior with reference to a Universal Turing machine $U$. Since there are infinitely many Universal Turing machines, there are infinitely many Solomonoff priors. Furthermore, these priors will often disagree in their verdicts. How much of a problem is this? Let us take a closer look.

Consider our example above. Suppose you are given the initial segment of a binary string:
\begin{equation*}
0000000000...
\end{equation*}
Given this initial segment, what is your prediction for the next bit? 

You might hope that Solomonoff prediction can vindicate the intuitive verdict that the next bit is likely to be a zero. There is an intuitive sense in which a string consisting entirely of zeros is `simple', and you might hope that our formal framework captures this intuition, because the shortest program which outputs a string of all zeros is shorter than the shortest program which outputs a string of ten zeros followed by ones. 

In Python, for example, one of the shortest programs to output a string of all zeros might be the following:
\begin{python}
while True:
    print(0)
\end{python}
In contrast, one of the shortest programs to output a string of ten zeros followed by ones might be the following more complicated program:
\begin{python}
i = 0
while True:
	while i <= 9:
	     print(0)
	     i = i + 1
	print(1)
\end{python}
Thus, it seems reasonable to expect that our Solomonoff predictor should assign a high probability to the next bit being zero.

If you find this kind of reasoning compelling, you might also hope that Solomonoff prediction helps us to handle the `New Riddle of Induction' and tells us why, after observing a number of green emeralds, we should predict that the next emerald is green rather than \emph{grue} (either green and already observed, or blue and not yet observed) \parencite{Goodman1955}.\footnote{See \cite{Elgin1997} for a collection of classic papers on the `New Riddle of Induction'.} Both the hypothesis that all emeralds are green and that all emeralds are grue fit our data equally well, but perhaps the all-green hypothesis is simpler and so should get a higher prior probability.\footnote{A similar line of argument is suggested by \textcite[42]{Vallinder2012}.}

However, such hopes are quickly disappointed. This is because different Universal Turing machines differ in how they measure the Kolmogorov complexity of strings. Relative to a `natural' Universal Turing machine, a string with all zeros is simpler than a string with some zeros first and ones after. However, relative to a `gruesome' Universal Turing Machine, a string with some zeros first and ones afterwards is simpler. If we think about the issue in terms of programming languages, this is quite obvious---it all depends on which operations in our programming language are taken to be primitive. Thus, different Solomonoff priors will license different predictions: Some will predict that a sequence of zeros will continue with a zero, others will predict that a sequence of zeros will continue with a one. Thus, if we use one of the Solomonoff priors, there is \emph{no guarantee whatsoever} that, after observing a long sequence of zeros, we assign a high probability to the next bit being zero.

The argument just sketched is a variant on the familiar point that simplicity is language dependent. Therefore, different choices of language (Universal Turing machine) will lead to different priors.\footnote{Readers familiar with \textcite{Goodman1955} will recognize that a version of this argument was leveled by Goodman against the idea that `green' is more simple than `grue'---it all depends on your choice of primitives.} Without a principled reason for why a `natural' Universal Turing machine should be preferred over a `gruesome' Universal Turing machine, the framework of Solomonoff prediction does not give us any reason for why, given an initial sequence of zeros, we should predict that the next bit is a zero rather than a one. Therefore, it does not look like the framework of Solomonoff prediction is any help in distinguishing `normal' and `gruesome' inductive behavior. As a consequence, it does not look like the framework of Solomonoff prediction gives a satisfying solution to the problem of the priors.

However, proponents of Solomonoff predictions can respond to this argument. According to them, the relativity of the Solomonoff prior to a choice of Universal Turing machine is not too worrying, because one can prove that all Solomonoff priors eventually \emph{converge} towards the same verdicts given more and more data. Thus, while different choices of Universal Turing Machine lead to different predictions in the short run, these differences `wash out' eventually.  So while there is an element of subjectivity in the choice of Universal Turing machine, this subjective element disappears in the limit. Call this the \emph{convergence reply}.\footnote{This reply is discussed by \textcite[1133]{Rathmanner2011}, \textcite[32]{Vallinder2012} and \textcite[473]{Sterkenburg2016}.}

Why is it true that different Solomonoff priors converge in their verdicts? To show this, we can invoke a standard convergence result from Bayesian statistics. To get this result on the table, we first need to introduce a bit more notation. Let $p$ and $p'$ be two probability measures on $\mathfrak{F}$. We define:
\begin{definition}
\emph{$p$ is absolutely continuous with respect to $p'$} if for all $A \in \mathfrak{F}$, \[ p(A) > 0 \implies p'(A) > 0.\]
\end{definition}
We now need a way of measuring the difference between two probability functions. Let $p$ and $p'$ be two probability functions on $\mathfrak{F}$. We define:
\begin{definition}
The \emph{total variational distance} between $p$ and $p'$ is \[sup_{A \in \mathfrak{F}} \mid p(A) - p'(A) \mid.\]
\end{definition}
\noindent Intuitively, the total variational distance between two probability functions defined on the same domain is the `maximal disagreement' between them.
We are interested in what happens after learning more and more data. To capture this, we define:
\begin{definition}
$E_n : \mathcal{B}^\infty \to \mathfrak{C}$ is the function which, given an infinite binary string $b \in \mathcal{B}^\infty$, outputs the cylinder set of strings which agree with $b$ in the first $n$ places.
\end{definition}
\noindent Intuitively, $E_n$ is a random variable which tells us the first $n$ digits of the string we are observing.\footnote{One can prove the Bayesian convergence result in a considerably more general setting, working with an abstract probability space and modeling evidence as sequence of increasingly fine-grained finite partitions (or sub $\sigma$-algebras). However, it is sufficient for our purposes to work with the measurable space $\langle \mathcal{B}^\infty, \mathfrak{F} \rangle$ introduced earlier.} We further define:
\begin{definition}
A probability function $p : \mathfrak{F} \to [0,1]$ is \emph{open-minded} if $p(\Gamma_x) > 0$ for all $x \in \mathcal{B}^*$.
\end{definition}
\noindent This captures the class of probability functions which do not rule out any finite initial sequence by assigning probability zero to it.

We want to talk about arbitrary probability functions $p : \mathfrak{F} \to [0,1]$, so we write $\Delta(\mathfrak{F})$ for the set of all probability functions on $\mathfrak{F}$. Now we define:
\begin{definition}
For any open-minded probability function $p : \mathfrak{F} \to [0,1]$, $p(\cdot \mid E_n) : \mathcal{B}^\infty \to \Delta(\mathfrak{F})$ is the function which outputs $p(\cdot \mid E_n(b))$ for each $b \in \mathcal{B}^\infty$.
\end{definition}
\noindent So $p(\cdot \mid E_n)$ is the result of conditionalizing $p(\cdot)$ on the first $n$ digits of the observed sequence. To make sure that $p(\cdot \mid E_n)$ is always well-defined, we restrict our attention to open-minded probability functions.

Now we can invoke the following well-known result in Bayesian statistics \parencite{BlackwellDubin1962}:\footnote{This and related results are discussed extensively by \cite{Earman1992}, \cite{Huttegger2015}, \cite{NielsenStewart2018}, \cite{NielsenSteward2019}.}
\begin{theorem}\label{thm:merge}
Let $p$ and $p'$ be two open-minded probability functions on $\mathfrak{F}$ such that $p$ is absolutely continuous with respect to $p'$. Then, we have
\begin{equation*}
\lim_{n \to \infty} sup_{A \in \mathfrak{F}} \mid p(A \mid E_n) - p'(A \mid E_n) \mid = 0,
\end{equation*}
$p$-almost surely. Therefore, $p$-almost surely, the total variational distance between $p$ and $p'$ goes to zero as $n \to \infty$.
\end{theorem}
Let me briefly comment on this result. First, to say that the equality holds `$p$-almost surely' means that it holds for all binary sequences except perhaps a set to which $p$ assigns probability zero. Second, as a direct corollary, if $p$ is absolutely continuous with respect to $p'$ and vice versa---so $p$ and $p'$ agree on which events have prior probability zero---then $p$ and $p'$ will also agree that, almost surely, their maximal disagreement will converge to zero as they observe more and more data. This captures a  natural sense of what it means for $p$ and $p'$ to converge in their verdicts.

With this result in place, the (almost sure) asymptotic equivalence of all Solomonoff priors follows straightforwardly.\footnote{For the purpose of stating the convergence result, I will assume that the Solomonoff priors are normalized to be probability measures on $\mathfrak{F}$. It is possible to obtain convergence result with the weaker assumption that Solomonoff priors are semi-measures, but there are difficulties in interpreting these results \parencite[200]{Sterkenburg2018thesis}---so to simplify our discussion, I'll stick with probability measures.}  Let $\lambda_{U}$ and $\lambda_{U'}$ be two Solomonoff priors defined relative to two Universal Turing Machines $U$ and $U'$. Now $\lambda_{U^\prime}$ is absolutely continuous with respect to $\lambda_{U}$ because $\lambda_U$ \emph{dominates} $\lambda_{U^\prime}$, which means that there is a constant $c$, depending on $U$ and $U'$, such that for all $x \in \mathcal{B}^*$, we have $\lambda_U(x) \geq c \lambda_{U'}(x)$  \parencite[71-2]{Sterkenburg2018thesis}. This is because the shortest programs producing a given string relative to two different Universal Turing machines cannot differ by more than a constant, as stated by the \emph{Invariance Theorem} \parencite[105]{Li2019}. Since $\lambda_{U}$ and $\lambda_{U'}$ were arbitrary, it follows that all Solomonoff priors are absolutely continuous with respect to each other.

Furthermore, each Solomonoff prior is open-minded. This is because it assigns positive probability to all computable sequences and every finite sequence is computable. (In the worst case, we can just hard-code the sequence into our program.) Therefore, by theorem \ref{thm:merge}, we have
\begin{equation*}
\lim_{n \to \infty} sup_{A \in \mathfrak{F}} \mid \lambda_U(A \mid E_n) - \lambda_{U'}(A \mid E_n) \mid = 0,
\end{equation*}
almost surely, so $\lambda_U$ and $\lambda_{U'}$ converge towards the same verdicts. Thus, all the infinitely many Solomonoff priors are (almost surely) asymptotically equivalent.\footnote{The `almost sure' qualification matters: it is \emph{not} true that different Solomonoff priors are asymptotically equivalent on \emph{all} sequences, as shown by \textcite[95]{Sterkenburg2018thesis} drawing on \textcite{Hutter2007semi}. However, this is generally true of Bayesian convergence theorems and no particular problem affecting Solomonoff prediction. For this reason, I will continue to say that different Solomonoff priors are `asymptotically equivalent' and sometimes drop the qualifier `almost surely'.}

As another consequence, we can show that any Solomonoff prior converges (almost surely) to optimal predictions on any sequence which is generated by some computable stochastic process \parencite[467]{Sterkenburg2016}. This means that we can think about the Solomonoff prior as a `universal pattern detector' which makes asymptotically optimal predictions on the minimal assumption that the data we are observing is generated by some computable process.

There is much more to say about the convergence reply. In particular, worries about subjectivity in the short run remain unaffected by long-run convergence results of the kind explained above \parencite[314]{Elga2016}. We still have no argument for why, after observing a finite number of green emeralds, it is more reasonable to predict that the next emerald is green rather than grue. However, I am happy to grant for the sake of argument that long-run convergence endows Solomonoff prediction with some kind of desirable objectivity. The focus of my argument is how the emphasis on long-run convergence interacts with another problematic feature of Solomonoff prediction: the fact that none of the Solomonoff priors are themselves computable.   

\section{Computability and Approximation}

There is a second problem for Solomonoff prediction: None of the infinitely many Solomonoff priors are computable. This means that there is no possible algorithm which will tell us, after finitely many steps, what the Solomonoff prior of a particular binary sequence \emph{is}---even if we have fixed a choice of Universal Turing machine.

Let us first define what it means for a pre-measure $p : \mathfrak{C} \to [0,1]$ to be computable, following \textcite[36]{Li2019}:
\begin{definition}
$p : \mathfrak{C} \to [0,1]$ is computable if there exists a computable function $g(x, k) : \mathfrak{C} \times \mathbb{N} \to \mathbb{Q}$ such that for any $\Gamma_x \in \mathfrak{C}$ and $k \in \mathbb{N}$, 
\begin{equation*}
\mid p(\Gamma_x) - g(\Gamma_x, k) \mid < \frac{1}{k}.
\end{equation*}
\end{definition}
This means that a pre-measure $p: \mathfrak{C} \to [0,1]$ is computable if there is an algorithm which we can use to approximate $p(\Gamma_x)$ to any desired degree of precision for any cylinder set $\Gamma_x \in \mathfrak{C}$. 

Then, we have the following:
\begin{theorem}
For any Universal Turing Machine $U$, $\lambda_{U}$ is not computable \parencite[303]{Li2019}.
\end{theorem}
\noindent \textcite{Leikehutter2018} discuss further results on the computability of Solomonoff prediction and related frameworks.

Since it seems plausible that we can only use computable inductive methods, this looks like a big problem. It is impossible for anyone to actually use Solomonoff prediction for inference or decision making. The lack of computability also seems to undermine the intended application of Solomonoff prediction as a foundation for artificial intelligence, since it is impossible to build an AI system which uses Solomonoff prediction. One might worry that for this reason, Solomonoff prediction is \emph{completely useless} as a practical guide for assigning prior probabilities. Further, the lack of computability might cut even deeper. It is unclear whether it is even possible for us, or any AI agent we might build, to `adopt' one of the uncomputable Solomonoff priors. I will return to this issue below.

Again, proponents of Solomonoff prediction can respond to this argument. While it is true that Solomonoff prediction is not computable, it is \emph{semi-computable}, which means that there are algorithms which get closer to $\lambda_{U}(x)$ at each step. This means that there are algorithms which \emph{approximate} the Solomonoff prior in some sense. Call this the \emph{approximation reply}.\footnote{This reply is discussed by \textcite[11]{Solomonoff1964} and \textcite[8-9]{Solomonoff2009}.}  

To see how such approximations could work, let me first explain in a bit more detail \emph{why} the Solomonoff prior is not computable. Recall that the Solomonoff prior of a binary string $b$ is inversely proportional to the Kolmogorov complexity of $b$: the length of the shortest program which outputs $b$, given some Universal Turing Machine. However, Kolmogorov complexity is not computable.\footnote{\textcite{Chaitin1995} provide a direct proof of this fact by reducing the problem of computing Kolmogorov complexity to the Halting problem.} There is no possible algorithm which, given an arbitrary binary string, outputs the Kolmogorov complexity of that string. As a consequence, the Solomonoff prior is not computable.

However, while Kolmogorov complexity is not computable, there are computable approximations to it. To simplify drastically, we can approximate the Kolmogorov complexity of a given string by stopping the search for the shortest program which outputs that string after a fixed time and consider the shortest program \emph{so far} which outputs the string. Call this \emph{bounded Kolmogorov complexity}.\footnote{See \cite[Chapter 7]{Li2019} for a rich discussion.} We can define a prior which assigns probability inversely proportional to bounded Kolmogorov complexity. As we let the search time go to infinity, we recover the original Kolmogorov complexity of our string.\footnote{\textcite{Veness2011} provide a concrete approximation to Solomonoff prediction. Also see \cite{Schmidhuber2002}.}

Given such approximations, one might hope that Solomonoff prediction is still a useful constraint on priors. It provides an ideal for the prior probabilities of a computationally unbounded reasoner, and in practice, we should do our best to approximate this ideal using our finite computational resources. This attitude is expressed, for example, when \textcite[83]{Solomonoff1997a} writes that ``despite its incomputability, algorithmic probability can serve as a kind of `gold standard' for induction systems''.

As before, there is much more to say about this argument, which raises interesting questions about `ideal theorizing' and the value of approximation.\footnote{See \cite{Staffel2019, CarrForthcoming} for recent discussions of `ideal' vs. `non-ideal' theorizing in epistemology and the value of approximation.} However, I am happy to grant for the sake of argument that there may be something valuable about an ideal theory which can never be implemented but only approximated.

There are some messy details which I'm ignoring here. First, it turns out that the Solomonoff \emph{predictor} is not even semi-computable \parencite[651]{Sterkenburg2019}. Furthermore, the normalized Solomonoff prior is not even semi-computable \parencite{Leikehutter2018}. Both only satisfy the weaker requirement of \emph{limit computability}: there is an algorithm which will converge to the correct probability value in the limit, but is \emph{not} guaranteed to get closer at each step. These messy details make it harder to maintain the convergence reply, because they make it harder to see how we could have \emph{any} sensible method for approximating Solomonoff prediction. However, the point I will discuss next is an \emph{additional} problem even if these messy details can somehow be cleaned up. 
 
\section{A Dilemma}

When pressed on the relativity of the Solomonoff prior to a Universal Turing machine, it is natural to appeal to asymptotic convergence. When pressed on the uncomputability of the Solomonoff prior, it is natural to appeal to computable approximations. However, there is a deep tension between the convergence reply and the approximation reply.

The tension arises for the following reason. Suppose we accept the approximation reply. We hold that while Solomonoff prediction is not computable, we can use some computable approximation of Solomonoff prediction to guide our inductive reasoning and construct AI systems. However, this response undercuts the convergence reply because, for reasons I will explain in a moment, \emph{different computable approximations to Solomonoff prediction are not necessarily asymptotically equivalent}. Therefore, we can no longer respond to the worry about language dependence by invoking long-run convergence.

To see why different computable approximations to Solomonoff prediction are not guaranteed to converge, recall first that  different Solomonoff priors \emph{do} converge because they are absolutely continuous with respect to each other. Now consider some computable approximation to Solomonoff prediction. There are different ways to spell out what it means to `approximate' the Solomonoff prior, but for my argument, the details of how we think about our `approximation strategy' will be largely irrelevant. As explained above, there are considerable difficulties in whether we can make sense of such an approximation strategy for the Solomonoff predictor and normalized Solomonoff prior, since they are only limit computable. I will sidestep these difficulties by treating the approximation strategy as a black box---what matters is just that our computable approximation to the Solomonoff prior is \emph{some computable probability measure}. 

Why should it be a probability measure, as opposed to a semi-measure? For standard Bayesian reasons: to avoid dutch books and  accuracy dominance. Why should it be computable? Because the whole point of the approximation reply is that we can actually use the approximation to make inferences and guide decisions. So we should better be able to compute, in a finite time, what the probability of a given event is. Otherwise, the approximation reply seems like a non-starter.

So let us consider some approximation to Solomonoff prediction, which is some computable probability measure. I claim that this computable approximation must assign probability zero to some computable sequence. This is because  \emph{every computable probability measure assigns probability zero to some computable sequence}:
\begin{theorem}
Let $p : \mathfrak{F} \to [0,1]$ be a computable probability measure. Then, there is some computable $b \in \mathcal{B}^\infty$ such that $p(b) = 0$.
\end{theorem}
This result is originally due to \textcite{Putnam1963}, who gives a beautiful `diagonal argument' for it.\footnote{For a wide-ranging discussion of Putnam's argument, see \cite[Chapter 9]{Earman1992}. In statistics, a similar result is due to \textcite{Oakes1985}, which is explicitly connected to Putnam's argument by \textcite{Dawid1985}. See also \cite{Schervish1985}.} Consider some computable prior $p$. Here is how to construct a `diagonal sequence' $D$ for our prior $p$, where $D_i$ denotes the $i$-th bit of $D$ and $E_n$ denotes the first $n$ bits of $D$:
\begin{equation*}
D_1 = 0
\end{equation*}
\begin{equation*}
    D_{n+1} =
    \begin{cases*}
      1 & if $p(1 \mid E_n) < \frac{1}{2}$ \\
      0 & if $p(1 \mid E_n) \geq \frac{1}{2}$
   	\end{cases*}
\end{equation*}
We arbitrarily start our sequence with a zero. To determine the next digit, we first check what our prior $p$ predicts after observing a zero. Then, we do the opposite. We iterate this procedure infinitely many times, and our binary sequence $D$ is finished. Since we have assumed that $p$ is computable, $D$ must be computable as well. 

Now why must $p$ assign probability zero to $D$? Because by construction, $p(D_{n+1} \mid E_n)$ can never go above $\frac{1}{2}$. Therefore, even though the sequence we are observing is generated by a deterministic computable process, our computable prior cannot predict the next bit better than random guessing. However, if $p(D)$ were greater than zero, then $p(D_{n+1} \mid E_n)$ would eventually climb above $\frac{1}{2}$, which contradicts our assumption.

\textcite{Sterkenburg2019} discusses the relationship between Solomonoff prediction and Putnam's diagonal argument and concludes that ``Putnam's argument stands'' \parencite[653]{Sterkenburg2019}. In particular, Putnam's argument provides an alternative way to prove that the Solomonoff prior is not computable.\footnote{Further, \textcite[651]{Sterkenburg2019} points out that we can use Putnam's argument to show that the Solomonoff predictor is not semi-computable but only limit-computable.} My argument here is different, since my point is that we can use Putnam's argument to highlight a deep tension between the approximation reply and the convergence reply. While the tension between the approximation reply and the convergence reply is a relatively straightforward consequence of Putnam's diagonal argument, this particular point has not received any attention in the debate surrounding Solomonoff prediction. I conjecture that this is because the convergence reply and the approximation reply are often discussed separately, while not enough attention is paid to how they interact with each other. The convergence reply inhabits the realm of `ideal theorizing', where we don't really care about constraints of computability, while the approximation reply tries to connect ideal theory to the real world. However, it is important to pay close attention to how these different features of our theory interact. With this paper, I hope to take some steps to remedy this `cognitive fragmentation'.

After clarifying what this paper aims to accomplish, let's get into the argument. Suppose we use a computable approximation to Solomonoff prediction. \emph{The key point is that we face a choice between different approximations which are not guaranteed to be asymptotically equivalent.} 

Consider two different computable priors $p$ and $p'$ which approximate Solomonoff prediction in some sense. Note that this could mean two different things: it could mean that we fix a given Solomonoff prior $\lambda_U$ and use two different `approximation strategies'. Alternatively, it could mean that we fix an `approximation strategy' and apply it to two different Solomonoff priors $\lambda_U$ and $\lambda_{U'}$ based on different Universal Turing Machines. The second possibility is closely related to the kind of language dependence discussed earlier---we might face the choice between a `natural' and a `gruesome' Universal Turing Machine. The first possibility seems a bit different, it is best characterized as a kind of `approximation dependence'. My argument will work with either of these options.

So we have two computable approximations $p$ and $p'$. This means, as I have argued above, that both $p$ and $p'$ are computable probability measures. By Putnam's argument, both $p$ and $p'$ assign probability zero to some computable sequences. Call these sequences $D$ and $D'$. Note, first, that both $p$ and $p'$ rule out some computable hypotheses and so seem to make substantive assumptions about the world \emph{beyond} computability. For those who hold that Solomonoff prediction gives us an `universal pattern detector' which can find any computable pattern, this is already a problem, because the approximations $p$ and $p'$ cannot find \emph{every} computable pattern. This is a first hint that the asymptotic properties which make Solomonoff prediction great are \emph{not} preserved in computable approximations to Solomonoff prediction.

Now the key point for my argument is that, if $p$ and $p'$ are different, then $D$ and $D'$ might be different as well. So $p$ might assign a positive probability to $D'$. Conversely, $p'$ might assign a positive probability to $D$. The crucial observation is that while each prior $p$ is forced to assign probability zero to its `own' diagonal sequence $D$ on pain of inconsistency, no inconsistency arises when some prior $p$ assigns positive probability to the diagonal sequence $D'$ for some \emph{other} prior $p'$.\footnote{Here is a simple example. Let $p'$ be generated by the uniform measure which assigns probability $2^{-n}$ to each binary sequence of length $n$. Applying Putnam's construction, the diagonal sequence $D'$ for this prior is the sequence $s_0$ consisting of all zeros. However, we can easily find \emph{another} (computable) prior $p$ which assigns positive probability to $s_0$, just let $p(\{s_0\}) = 1$.}

In the case just discussed, $p$ and $p'$ fail to be absolutely continuous with respect to each other, since they differ in what events are assigned  probability zero. Therefore, it is \emph{not} guaranteed that $p$ and $p'$ are (almost surely) asymptotically equivalent. They might yield different verdicts forever. This means that if there is a subjective element in the choice between $p$ and $p'$, this subjective element is \emph{not} guaranteed to `wash out' in the long run.

To bring this out more clearly, we can draw on a recent result by \textcite{NielsenStewart2018}. They relax the assumption of absolute continuity and study what happens to Bayesian convergence results in this more general setting. What they show is the following: If prior $p$ is \emph{not} absolutely continuous with respect to prior $p'$, then $p$ must assign some positive probability to the event that $p$ and $p'$ \emph{polarize}, which means that the total variational distance between them converges to 1 as they learn an increasing sequence of shared evidence.\footnote{See their theorem 3, which generalizes the classic merging-of-opinion results due to \textcite{BlackwellDubin1962}.} So if two priors fail to be absolutely continuous with respect to each other, they must assign positive probability to the event that learning shared evidence drives them towards maximal disagreement.

I have argued above that two computable approximations of the Solomonoff prior might fail to be absolutely continuous with respect to each other. In combination with the result by \textcite{NielsenStewart2018}, this means that two computable approximations of the Solomonoff prior might assign positive probability to polarization in the limit: further evidence drives them towards maximal disagreement. This gives us a clear sense in which, when we consider computable approximations to the Solomonoff prior, subjectivity is \emph{not} guaranteed to `wash out' as we observe more evidence.  This, in turn, means that the choice between our two approximations introduces a significant subjective element which is \emph{not} guaranteed to wash out, but might, with positive probability, persist indefinitely. This looks like bad news for the convergence reply.

Let me add an important clarification. My argument shows that for two computable approximations $p$ and $p'$ of the Solomonoff prior, it is not guaranteed that that $p$ and $p'$ will converge \emph{without making further assumptions}. We might add additional requirements on `acceptable approximations' which rule out such cases by forcing all computable approximations to the Solomonoff prior to be absolutely continuous with respect to each other. However, any such strategy faces a deep problem. Since each computable prior must assign probability zero to some computable sequence, this would mean that our set of approximations to the Solomonoff prior rules out some computable sequences \emph{a priori}. However, this looks incompatible with the motivation behind Solomonoff prediction. The Solomonoff prior is supposed to be a `universal pattern detector' which can learn any computable pattern. So the price for forcing asymptotic agreement among different approximations to the Solomonoff prior would be to make substantive assumptions \emph{beyond} computability, which is exactly what Solomonoff prediction was designed to avoid. 

So there is a deep tension between the convergence reply and the approximation reply. If we accept the approximation reply, this means that we should use some computable approximation to the Solomonoff prior to guide our inductive reasoning. However, the move to computable approximations undercuts the convergence reply, since different computable approximations are \emph{not} necessarily asymptotically equivalent. They might, with positive probability, yield different verdicts forever, and \emph{never} converge to the same predictions. Therefore, we can no longer dismiss the worry about language dependence by invoking long-run convergence. For example, if two different approximations arise from two different Universal Turing Machines, the difference between `natural' and `gruesome' Universal Turing Machines is \emph{not} guaranteed to wash out in the long run, but might stay with us forever. So we better come up with some good reasons for why we should use a `natural' rather than a `gruesome' Universal Turing machine.\footnote{See, for example, \textcite[1113]{Rathmanner2011}, who (inconclusively) explore the issue of whether some Universal Turing Machines might be more `natural' than others.} More generally, we have to face the problem of subjectivity in the choice of Universal Turing machine head-on and cannot downplay the significance of this choice by invoking asymptotic convergence. In fact, the situation is even more bleak: Even if we find convincing arguments for why some Universal Turing Machine is the `correct' or `natural' one, we might still face the choice between different `approximation strategies' which introduce a persistent subjective element. So when we consider computable approximations to Solomonoff prediction, both language dependence and approximation dependence introduce subjective elements which are \emph{not} guaranteed to wash out.

Suppose, on the other hand, that we are convinced by the convergence reply. In this case, we think that what makes Solomonoff prediction great is that different choices of Universal Turing machine lead to priors which are (almost surely) asymptotically equivalent and which assign positive probability to all computable sequences. However, in this case we have to embrace that Solomonoff prediction is essentially uncomputable. This is because there is no computable prior which assigns positive probability to all computable sequences. So the emphasis on convergence undercuts the approximation reply. From this perspective, what makes Solomonoff prediction great is its asymptotic behavior. \emph{However, no computable approximation to Solomonoff prediction preserves this great asymptotic behavior.} Therefore, it is not clear why there is any point in using a computable approximation to Solomonoff prediction to guide our inductive inferences or as a foundation for AI. 

You might object to my argument as follows: `Suppose I adopt the Solomonoff prior. In response to the charge that it's not objective, I invoke convergence. In response to the charge that the Solomonoff prior is not computable, I invoke approximation. In response to the charge that these computable approximations need not themselves converge, I simply deny that there's any problem. The computable approximations are not \emph{my probabilities}, they are just useful computational tools that I can use to calculate and report my (approximate) probabilities.'\footnote{Thanks to an anonymous referee for pressing this objection.}

Let me reply to this objection by making clear what the target of my argument is. I grant that if one can really `adopt' one of the Solomonoff priors and use computable approximations merely as a tool to report one's probabilities, this gets around the problem. But is it really possible for us, or an AI agent we build, to adopt an incomputable probability function as a prior? This depends on what makes it the case that an agent has a particular prior, which is a difficult question I cannot fully discuss here. But it seems plausible that any physically implemented agent can only represent and act according to a computable prior. Therefore, it is unclear whether we can really `adopt' an uncomputable prior. The same reasoning holds for any AI system which we might construct. The best we can do is to adopt some approximation to the Solomonoff prior, and my point is that we face some difficult choices in choosing such an approximation.

\section{Convergence for Subjective Bayesians}

Let me finish by briefly discussing how my argument relates to broader questions in Bayesian epistemology. As we have seen in the beginning, one of the big questions for Bayesians is how to choose a prior---the problem of the priors. Solomonoff prediction is an attempt to solve this problem by specifying a `universal' prior. But, as I have argued, this ambition ultimately fails, because we lose guaranteed convergence if we use computable approximations to the Solomonoff prior. 

One might wonder whether this argument poses problems for Bayesian convergence arguments more generally. Bayesians often argue that the choice of prior is not very significant, because given `mild' assumptions, different priors converge as more data is observed.\footnote{See, for example, the classic discussion in \cite[Chapter 6]{Earman1992}.} However, the key assumption is absolute continuity: different priors must assign positive probability to the same events. And Putnam's argument shows that every computable prior must assign probability zero to some computable hypothesis. Taken together, this suggests that we can only hope for convergence if we agree on substantive assumptions about the world---beyond computability. So the scope of Bayesian convergence arguments is more limited than one might have hoped.\footnote{This is also the conclusion of \textcite{NielsenStewart2018}, who argue that Bayesian rationality is compatible with persistent disagreement after learning shared evidence.}

This should not come as a surprise to \emph{subjective} Bayesians who hold that the choice of prior embodies substantive assumptions which reflect the personal beliefs of an agent. Consider, for example, the following passage in \textcite{Savage1972} defending a `personalistic' (subjective Bayesian) view of probability: ``The criteria incorporated in the personalistic view do not guarantee agreement on all questions among all honest and freely communicating people, even in principle. That incompleteness, if one will call it such, does not distress me, for I think that at least some of the disagreement we see around us is due neither to dishonesty, to errors in reasoning, nor to friction in communication [...]'' \parencite[67-8]{Savage1972}.

If you agree that the choice of prior embodies a subjective element, then the fact that we cannot guarantee convergence without shared substantive assumptions should not come as a shock. So my argument does not raise new problems for subjective Bayesians. However, it raises problems for any attempt to define a `universal' or `objective' prior which does not embody substantive assumptions about the world.

\section{Conclusion}

Proponents of Solomonoff prediction face a dilemma. They cannot simultaneously respond to worries about language dependence by invoking asymptotic convergence while responding to worries about uncomputability by invoking computable approximations. This is because, for very general reasons, no computable approximation to Solomonoff prediction has the same asymptotic behavior as the Solomonoff priors. 

In the absence of principled criteria for choosing a Universal Turing machine, it looks like Solomonoff prediction is either subject to thorny problems of subjectivity and language dependence, or else essentially uncomputable and therefore useless as a guide to scientific inference and the design of optimal artificial agents.

\printbibliography

@article{BlackwellDubin1962,
	year = {1962},
	journal = {The Annals of Mathematical Statistics},
	number = {3},
	title = {Merging of Opinions with Increasing Information},
	doi = {10.1214/aoms/1177704456},
	pages = {882-86},
	volume = {33},
	author = {David Blackwell and Lester Dubin}
}

@article{CarrForthcoming,
	pages = {1-32},
	title = {Why Ideal Epistemology?},
	doi = {10.1093/mind/fzab023},
	year = {forthcoming},
	author = {Jennifer Rose Carr},
	journal = {Mind}
}

@techreport{Chaitin1995,
  	title={Program-Size Complexity Computes the Halting Problem},
  	author={Chaitin, Gregory J and Arslanov, Asat and Calude, Cristian},
  	year={1995},
	type = {Technical Report},
  	number={CDMTCS-008},
  	institution={Department of Computer Science, The University of Auckland, New Zealand}
}

@article{Chater2013,
	author = {Fulop, Sean and Chater, Nick},
	title = {Editors' Introduction: Why Formal Learning Theory Matters for Cognitive Science},
	journal = {Topics in Cognitive Science},
	volume = {5},
	number = {1},
	pages = {3-12},
	doi = {https://doi.org/10.1111/tops.12004},
	year = {2013}
}

@article{DeFinetti1937, 
	author = {Bruno de Finetti}, 
	journal = {Annales de l'Institut Henri Poincar\'e}, 
	pages = {1--68}, 
	title = {La Pr\'evision: Ses Lois Logiques, Ses Sources Subjectives}, 
	year = {1937}, 
	volume = {17} 
}

@article{Dawid1985, 
	title={Comment: The Impossibility of Inductive Inference}, 
	author={Dawid, A Philip}, 
	journal={Journal of the American Statistical Association}, 
	volume={80}, 
	number={390}, 
	pages={340-41}, 
	year={1985}, 
	doi={10.1080/01621459.1985.10478118} 
}

@book{Earman1992,
	year = {1992},
	author = {John Earman},
	title = {Bayes or Bust? A Critical Examination of Bayesian Confirmation Theory},
	publisher = {MIT Press}
}

@article{Elga2016, 
	volume = {83}, 
	number = {3}, 
	author = {Adam Elga}, 
	title = {Bayesian Humility}, 
	journal = {Philosophy of Science}, 
	doi = {10.1086/685740}, 
	year = {2016}, 
	pages = {305-23} 
}

@book{Elgin1997,
 	title={Nelson Goodman's New Riddle of Induction},
 	author={Elgin, Catherine Z},
 	series={The Philosophy of Nelson Goodman},
 	volume={2},
 	year={1997},
 	publisher={Taylor \& Francis}
}

@book{Goodman1955,
	year = {1955},
	author = {Nelson Goodman},
	title = {Fact, Fiction, and Forecast},
	publisher = {Harvard University Press}
}

@article{Huttegger2015,
	number = {4},
	title = {Merging of Opinions and Probability Kinematics},
	volume = {8},
	journal = {Review of Symbolic Logic},
	year = {2015},
	doi = {10.1017/s1755020315000180},
	author = {Simon M. Huttegger},
	pages = {611-48}
}

@article{Hutter2007,
	year = {2007},
	journal = {Theoretical Computer Science},
	number = {1},
	title = {On Universal Prediction and Bayesian Confirmation},
	doi = {10.1016/j.tcs.2007.05.016},
	pages = {33-48},
	volume = {384},
	author = {Marcus Hutter}
}

@article{Hutter2007semi,
	title={On Semimeasures Predicting Martin-L{\"o}f Random Sequences},
	author={Hutter, Marcus and Muchnik, Andrej},
	journal={Theoretical Computer Science},
	volume={382},
	number={3},
	pages={247-61},
	year={2007},
	doi={10.1016/j.tcs.2007.03.040}
}

@book{Hutter2004, 
	title={Universal Artificial Intelligence: Sequential Decisions Based on Algorithmic Probability}, 
	author={Hutter, Marcus}, 
	year={2004}, 
	publisher={Springer},
	series={Texts in Theoretical Computer Science}, 
	doi={10.1007/b138233} 
}

@incollection{Icard2017,
	title={Beyond Almost-Sure Termination},
	author={Icard, Thomas},
	booktitle={Proceedings of the 39th Annual Meeting of the Cognitive Science Society},
	year={2017},
	editor={Gunzelmann, Glenn and Andrew Howes and Thora Tenbrink and Eddy Davelaar},
	place={Austin, TX: Cognitive Science Society},
	pages = {2255-60}
}

@article{KelleyForthcoming,
	journal = {Philosophy of Science},
	year = {forthcoming},
	doi = {10.1017/psa.2021.37},
	title = {On Accuracy and Coherence with Infinite Opinion Sets},
	author = {Mikayla Kelley}
}

@article{leikehutter2018, 
	title={On the Computability of Solomonoff Induction and AIXI}, 
	author={Leike, Jan and Hutter, Marcus}, 
	journal={Theoretical Computer Science}, 
	volume={716}, 
	pages={28-49}, 
	year={2018},
	doi={10.1016/j.tcs.2017.11.020} 
}

@book{Li2019, 
	title={An Introduction to Kolmogorov Complexity and Its Applications}, 
	author={Li, Ming and Vit{\'a}nyi, Paul}, 
	volume={4}, 
	publisher={Springer},
	year={2019}, 
	doi={10.1007/978-3-030-11298-1} 
}

@article{NielsenStewart2018,
	author = {Nielsen, Michael and Stewart, Rush},
	year = {2018},
	pages = {51-78},
	title = {Persistent Disagreement and Polarization in a Bayesian Setting},
	volume = {72},
	journal = {The British Journal for the Philosophy of Science},
	doi = {10.1093/bjps/axy056}
}

@article{NielsenSteward2019,
	author = {Nielsen, Michael and Stewart, Rush},
	title = {Another Approach to Consensus and Maximally Informed Opinions with Increasing Evidence},
	journal = {Philosophy of Science},
	volume = {86},
	number = {2},
	pages = {236-54},
	year = {2019},
	doi = {10.1086/701954},
}

@incollection{Putnam1963,
	editor = {Paul Arthur Schilpp},
	booktitle = {The Philosophy of Rudolf Carnap},
	publisher = {Open Court: La Salle},
	title = {Degree of Confirmation and Inductive Logic},
	pages = {761-783},
	author = {Hilary Putnam},
	year = {1963}
}

@book{Pettigrew2016,
	publisher = {Oxford University Press UK},
	year = {2016},
	author = {Richard Pettigrew},
	title = {Accuracy and the Laws of Credence}
}

@article{Predd2009,
	number = {10},
	title = {Probabilistic Coherence and Proper Scoring Rules},
	author = {Joel Predd and Robert Seiringer and Elliott Lieb and Daniel Osherson and H. Vincent Poor and Sanjeev Kulkarni},
	volume = {55},
	pages = {4786-92},
	journal = {IEEE Transactions on Information Theory},
	year = {2009},
	doi={10.1109/TIT.2009.2027573}
}

@incollection{Ortner2009,
	editor = {Dov M. Gabby and Stephan Hartmann and John Woods},
	booktitle = {Handbook of the History of Logic: Inductive Logic},
	author = {Ronald Ortner and Hannes Leitgeb},
	publisher = {Elsevier: Amsterdam},
	pages = {719-72},
	title = {Mechanizing Induction},
	year = {2009},
	doi = {10.1016/b978-0-444-52936-7.50018-5}
}

@article{Oakes1985,
	title={Self-Calibrating Priors Do Not Exist},
	author={Oakes, David},
	journal={Journal of the American Statistical Association},
	volume={80},
	number={390},
	pages={339},
	year={1985},
	doi={10.1080/01621459.1985.10478117}
}

@article{Rathmanner2011,
 	title={A Philosophical Treatise of Universal Induction},
 	author={Rathmanner, Samuel and Hutter, Marcus},
 	journal={Entropy},
 	volume={13},
	number={6},
	pages={1076-136},
	year={2011},
  	doi={10.3390/e13061076}
}

@incollection{Schmidhuber2002,
	year = {2002},
	booktitle = {Computational Learning Theory: Proceedings of COLT 2002},
	title = {The Speed Prior: A New Simplicity Measure Yielding Near-Optimal Computable Predictions},
	doi = {10.1007/3-540-45435-7_15},
	pages = {216-28},
	author = {J{\"u}rgen Schmidhuber},
	editor = {Jyrki Kivinen and Robert H. Sloan}
}

@article{Solomonoff1964,
	year = {1964},
	journal = {Information and Control},
	volume = {7},
	number = {1},
	title = {A Formal Theory of Inductive Inference. Part I},
	doi = {10.1016/S0019-9958(64)90223-2},
	pages = {1-22},
	author={Solomonoff, Ray J}, 
}

@article{Solomonoff1997a, 
	title={The Discovery of Algorithmic Probability}, 
	author={Solomonoff, Ray J}, 
	journal={Journal of Computer and System Sciences}, 
	volume={55}, 
	number={1}, 
	pages={73--88}, 
	year={1997}, 
	doi={10.1006/jcss.1997.1500} 
}

@incollection{Solomonoff2009,
	editor = {Frank Emmert-Streib and Matthias Dehmer},
	booktitle = {Information Theory and Statistical Learning,},
	publisher = {Springer},
	title = {Algorithmic Probability: Theory and Applications},
	pages = {1--23},
	author={Solomonoff, Ray J}, 
	year = {2009}
}

@book{Savage1972,
	year = {1972},
	publisher = {Wiley Publications in Statistics},
	title = {The Foundations of Statistics},
	author = {Leonard J. Savage},
	edition = {Second Revised Edition}
}

@article{Schervish1985,
	title={Self-Calibrating Priors Do Not Exist: Comment},
	author={Mark J Schervish},
	journal={Journal of the American Statistical Association},
	volume={80},
	number={390},
	pages={341--342},
	year={1985},
	doi={10.2307/2287893}
}

@book{Staffel2019,
	title = {Unsettled Thoughts: A Theory of Degrees of Rationality},
	publisher = {Oxford University Press},
	author = {Julia Staffel},
	year = {2019}
}

@article{Sterkenburg2016,
	year = {2016},
	journal = {Philosophy of Science},
	number = {4},
	publisher = {University of Chicago Press},
	title = {Solomonoff Prediction and Occam's Razor},
	doi = {10.1086/687257},
	pages = {459-79},
	volume = {83},
	author = {Tom F. Sterkenburg}
}

@phdthesis{Sterkenburg2018thesis, 
	title={Universal Prediction: A Philosophical Investigation}, 
	author={Sterkenburg, Tom F.},
	institution={Rijksuniversiteit Groningen}, 
	year={2018} 
}

@article{Sterkenburg2019,
	doi = {10.1007/s10670-018-9975-x},
	author = {Tom F. Sterkenburg},
	publisher = {Springer Verlag},
	volume = {84},
	title = {Putnam's Diagonal Argument and the Impossibility of a Universal Learning Machine},
	pages = {633-56},
	year = {2019},
	number = {3},
	journal = {Erkenntnis}
}

@thesis{Vallinder2012,
  	title={Solomonoff Induction: A Solution to the Problem of the Priors?},
  	author={Vallinder, Aron},
 	type={MA thesis},
  	institution={Lund University},
  	year={2012}
}

@article{Veness2011,
	title={A Monte-Carlo AIXI Approximation},
	author={Veness, Joel and Ng, Kee Siong and Hutter, Marcus and Uther, William and Silver, David},
	journal={Journal of Artificial Intelligence Research},
	volume={40},
	pages={95-142},
	year={2011},
	doi = {10.1613/jair.3125 }
}

@incollection{Wood2013, 
	title={(Non-)Equivalence of Universal Priors}, 
	author={Wood, Ian and Sunehag, Peter and Hutter, Marcus}, 
	booktitle={Algorithmic Probability and Friends. Bayesian Prediction and Artificial Intelligence}, 
	pages={417-25}, 
	year={2013}, 
	publisher={Springer},
	doi={10.1007/978-3-642-44958-1_33} 
}

\end{document}